……………………………

# Cybertrust: From Explainable to Actionable and Interpretable AI (AI$^2$)

Alexander Kott[1], Stephanie Galaitsi[2], Benjamin D. Trump[2], Jeffrey M. Keisler[3], Igor Linkov[2]

[1] US Army Futures Command, Adelphi, MD

[2] US Army Engineer Research and Development Center, Concord, MA

[3] University of Massachusetts, Boston, MA

## Abstract

To benefit from AI advances, users and operators of AI systems must have reason to trust it. Trust arises from multiple interactions, where predictable and desirable behavior is reinforced over time. Providing the system's users with some understanding of AI operations can support predictability, but forcing AI to explain itself risks constraining AI capabilities to only those reconcilable with human cognition. We argue that AI systems should be designed with features that build trust by bringing decision-analytic perspectives and formal tools into AI. Instead of trying to achieve explainable AI, we should develop interpretable and actionable AI. Actionable and Interpretable AI (AI$^2$) will incorporate explicit quantifications and visualizations of user confidence in AI recommendations. In doing so, it will allow examining and testing of AI system predictions to establish a basis for trust in the systems' decision making and ensure broad benefits from deploying and advancing its computational capabilities.

## Decision Making: Humans and Artificial Intelligence (AI)

*"Can I trust the recommendation of an AI agent?"* This question is difficult to answer, especially if the decision at stake is complex and spans different spatial and temporal scales. Such difficulty is exacerbated when the outcomes of an AI-influenced decision may heighten existing risks to humans or introduce new risks altogether. Yet such high-stakes situations have become routine within the diverse systems that currently incorporate AI, like controls for chemical plants, defense systems, and health insurance rate determinations. Stakeholders must be prepared not only to configure AI and its enabling technologies for a given industry or activity, but also to have tools and methodologies to examine and recognize its failures, limitations, and needs for quality control at various stages of its development and implementation.

The ultimate goal of AI is to provide users with actionable recommendations that meet both implicit and explicit goals of the decision makers and stakeholders. Recommendations generated from AI-based

……………………………

approaches hold advantages over human decision makers through their ability to analyze vast bodies of information in limited time in an objective and logic-centered fashion. In many situations, these benefits are clear and already implemented in practice, such as machine learning systems for the detection of phishing attempts [Khonji et al. 2013]. AI applications are also capable of providing multistep and adaptable strategies, as demonstrated by programs that compete in chess or Go, as well as AI-based cybersecurity systems [Al-Shaer et al. 2019; Kott et al. 2019]

However, AI recommendations may not account for decision maker values or specific mission needs. For example, following a cyberattack, an AI-generated decision engine may recommend disabling an application on the compromised computer system. Such an action may neutralize the threat posed by the compromised system, but could simultaneous endanger a mission, negatively impact a critical user, or enable the adversary to extend the duration or scope of the cyberattack. The broader scale impacts of the recommended path forward may not have been incorporated into the AI's design or scope, causing the AI decision processes to omit critical conditions that a human operator would implicitly account for. Such incomplete scoping of AI-driven analysis is especially problematic when unspoken, unacknowledged, or subjective variables influence or shape what a successful outcome looks like. For example, an AI system may not account for the need for a particular asset to be available to achieve a mission later on, or for psychological impact on the system's users. The AI solves the problem it is given, but it the human's role to ensure the recommendation's suitability in context. Similarly, the human users making this judgment will benefit from understanding the factors that produced the AI's decision, especially when that understanding helps the users see the value of factors they themselves could have overlooked.

While AI-driven analysis enhances our decision making ability, providing insight into AI's shifts in its analysis of needs, expectations, and mission requirements will ensure its decisions' relevance and credibility and make its expectations for the future explicit. If AI's analytical outputs do not account for these and other broader and potentially subjective concerns, an overly myopic focus upon a tactical decision can derail strategic mission requirements. As such, a more effective deployment of AI into decision making must resolve the 'black box' concerns of AI – in that it is unclear how to explain, interpret, or act upon AI's conclusions as its underlying algorithm and parameters are difficult to decipher.

## Inevitable Disagreements: The Challenge of Fusing Human and AI Decision Making Capabilities

We can expect AI recommendations to differ, in some percentage of circumstances, from the choice the operator alone would have made. There are three possibilities on how this plays out for any yes/no decision: the AI is more risk-averse than the human, the AI is more risk-tolerant than the human, or the AI and the human agree (Table 1).

|  |  | AI | |
|---|---|---|---|
|  |  | *Yes* | *No* |
| **Human** | *Yes* | Agreement | Disagreement |
|  | *No* | Disagreement | Agreement |

Table 1. Typology of Human-AI Assessment of Decision Strategy

……………………………

Assuming that the AI is correct more often than a human under the same time and resource constraints (underscoring the utility of AI applications), on average, the human who disagrees with the AI should still follow the AI's recommendations. The challenge of that moment of discordance, then, is to convince humans to trust the AI output despite their own opposing judgment.

Already there are situations in which trust is fragile between AI and human users: the term "techlash" refers to the growing animus towards technology, especially information technology. Techlash is a distrust that the technologies have users' best interests at heart, given some questionable behavior on the part of the companies that design and promote them.[Finn and DuPont 2020] If the benefits of superior AI decision making are to be realized and further developed, there must be avenues for users to establish a foundation for trust in a given AI agent's decisions (Siau and Wang, 2018).

Trust in social situations grows based on performance over time (Lewicki, and Wiethoff, 2000), and trust in AI can be developed the same way. But both objectives and situations are liable to change in time and space, and a static decision made under specific circumstances may have limited utility in divergent futures. The contemporary world changes quickly and sometimes dramatically, and AI decisions must be contextualized within a changing and uncertain threat space.

A decision is made in a moment of time, but the threat space that ultimately arises will determine its value in application. A decision that addresses a narrow range of futures may be less applicable than a robust decision that would provide value in a broad range of futures. However, if one future looks much more probable than all others, a decision tailored to its unique outcomes may provide the most benefits in the future.

## Explainable AI

The inability of AI algorithms to articulate the reasons for specific predictions and recommendations has increased calls for AI output formats designed to build human confidence and trust. Various initiatives aim to produce AI models that are more easily understood without overly sacrificing accuracy in predictions. For example, the DARPA's eXplainable AI program will build algorithms with "the ability to explain their rationale, characterize their strengths and weaknesses, and convey an understanding of how they will behave in the future." The initiative aims to enable users to understand and characterize the weaknesses of these models, such as programming biases, as well as appreciate the value-added for applying the AI to specific circumstances. Explainability refers to the extent that an algorithm's internal mechanics can be explained in human terms.

There are several limitations of explainable AI in its current framing. First, while there may be situations in which AI can be explained, some processes, in spite of their practical benefits, are too complex for human cognition. AI algorithms may not necessarily reflect the specific problem that humans are trying to solve, and providing more information that may not be relevant to the decision context may not increase levels of trust. However, it is still possible to render some relationships more transparent: in image processing, saliency methods use digital neural networks to provide maps according to pixel relevance within the image. The true fidelity of various methods can be difficult to measure (Tomsett et al. 2019), but their application promotes the idea that some relationships between inputs and outputs

……………………………

can be held consistent for both human and AI cognition operations. This, however, may leave much of the algorithmic processes unexplained.

Second, a key advantage of AI may well be its ability to avoid human-like behavior when that behavior is not actually optimal: AI can provide innovative strategies for achieving objectives deduced from the framed problem. In some cases, an AI game-playing agent triumphs over human rivals not because it improves upon known human strategies, but precisely because it deviates from those strategies. However, mandating that AI explain something that is counterintuitive to human operators may not help in trust-building. AI arrives at decisions through convoluted and complex algorithms (the "black box") that are generally shrouded from or impenetrable to human operators. Inviting humans into the box may jeopardize AI's true power by forcing it to conform to human recognition.

Yet human understanding (and, typically, acceptance) is predicated on AI conformity to recognizable cognition processes. Just because humans do not see the reason for a process does not make it inconsequential. AI may arrive at decisions by avenues that are unfamiliar or obtuse compared to those upon which humans have historically relied. Truly benefitting from AI may entail excusing it from the onus to explain itself to humans because such a demand constrains AI to the same values and limitations that have always underpinned human decisions.

## Interpretable AI

The transition from explainability to interpretability means moving from providing a reason for a decision to assessing meaning in the context of specific decision or mission needs. Like Explainable AI, Interpretable AI recognizes the tradeoff between transparency and accuracy enabled by computational power. Rather than seeking to optimize both, interpretable AI emphasizes understanding cause and effect within the AI system (Chafika and Taleb, 2020). Users can examine the sensitivities of the output recommendations to changes in the parameter inputs without needing to understand the algorithms' complex parameter computations. Interpretable AI should allow users to toggle the parameters that are most uncertain about studying the impacts of their changes, as well as to test the AI's reaction to changes against the users' own beliefs about underlying relationships between inputs and outputs.

For example, in determining which car a person should purchase, income should be an important factor. Within a system for interpretable AI, income relevance and effect could be verified by varying the income input within the model and viewing the subsequent changes in model output. If an AI system exhibited extreme sensitivity to income and little sensitivity to the difference between a three-person family and four-person family, the user could conclude that the AI system reflects at least some of the factors that the user deems most important in car selection.

Instead of explaining the AI results to humans, interpretable AI models allow users to place AI recommendations in the context of the decision problem. Interpretation does not imply that operators must understand the process of driving the AI recommendations. To this end, forging meaning, more than explanation, allows the AI to build the functionality around accuracy and complexity while ensuring that humans can find sufficient meaning in the outcomes to implement them.

……………………………

## Actionable AI

Ultimately, the foundations of decision maker actions are grounded upon evidence-based data (including AI recommendations), as well as strategic and tactical considerations which include decision context, mission needs, available resources, etc. AI systems should provide level of confidence and sufficient information so that the decision maker can trust suggested course of action. Limitations in AI recommendations may not be clear to the user until the results are applied and evaluated and could potentially result in costly mistakes. Most AI systems are designed for specific context, but users, for lack of other options, may apply them to circumstances outside the design capabilities. This is especially important in situation where system perform under wide-scale threats thus changing both operational and decision environment outside of AI system performance range, a movable threat space.

Decision theory in general and specifically multi criteria decision analysis (MCDA, Linkov et al., 2020) provides a template for visualizing tradeoffs inherent in selecting one alterative over another and quantifying relative level of confidence that AI system is placing on its recommendations. MCDA approaches typically require as inputs scores across several dimensions associated with different management alternatives and outcomes reflective of objective evidence-based data and subjective weights relating to tradeoffs across these dimensions relevant to the mission and values. A basic but typical approach is to calculate the total value score for an alternative as a linear weighted sum of its scores across several criteria. These scores can be translated in utility functions or other metrics relevant to confidence in courses of action recommended by AI systems. Linking MCDA with Scenario Analysis (Tourki et al., 2013) allows integration of movable threat space to ensure AI decisions are applied in ways that will be most beneficial given the uncertainty of the future. The implication for AI is that there may be value in a layer which maps the content of generic explanations into the specific terms a rational human decision maker would use to infer that a course of action is appropriate, e.g,. in terms of the criteria which such a decision maker would use in the absence of AI.

## Conclusions

The validity or trustworthiness of decisions is predicated upon AI's analysis, and the uncertainty surrounding AI's decision making algorithms makes it difficult to understand which parameters were used to arrive at a conclusion (or, equally importantly, how those parameters were weighted for importance relative to one-another). Without addressing these concerns within AI's earliest stages of research and development, much of society may be unwilling or unable to incorporate AI to complement their operations and decision making needs. For those that do make use of AI, such stakeholders may encounter AI-driven guidance that is antithetical to their core values or mission requirements. This can cause users to reject AI's analysis in favor of human decision making abilities alone, or possibly to adopt the AI-driven conclusions to their own detriment for the longer-term future. Neither outcome is desirable.

However, this human-AI tension can be avoided by creating a more effective, ethical, and transparent process that fuses the decision making needs and abilities of both actors. To benefit society and ensure applicability, AI systems need a mechanism to build operator confidence in AI recommendations for these increasingly complex decision-making processes. To this end, we propose that many of the goals

……………………………

of Explainable AI can be realized with Actionable and Interpretable AI (AI2) without penalties for the AI computational power and accuracy.

Figure 1 illustrates that building trust in AI systems requires transferring meaning and relationships from one coherent system of understanding to another, from AI to human cognition. Explainable AI may be possible in some circumstances, but is inevitably couched within the context (threat environment) and the objectives and predetermined notions of the user (mission space). By rendering these more visible within the AI interface structure, the user can better access aspects of understanding even if the black box itself, meaning the actual decision making computations, cannot be fully explained within human cognition constraints.

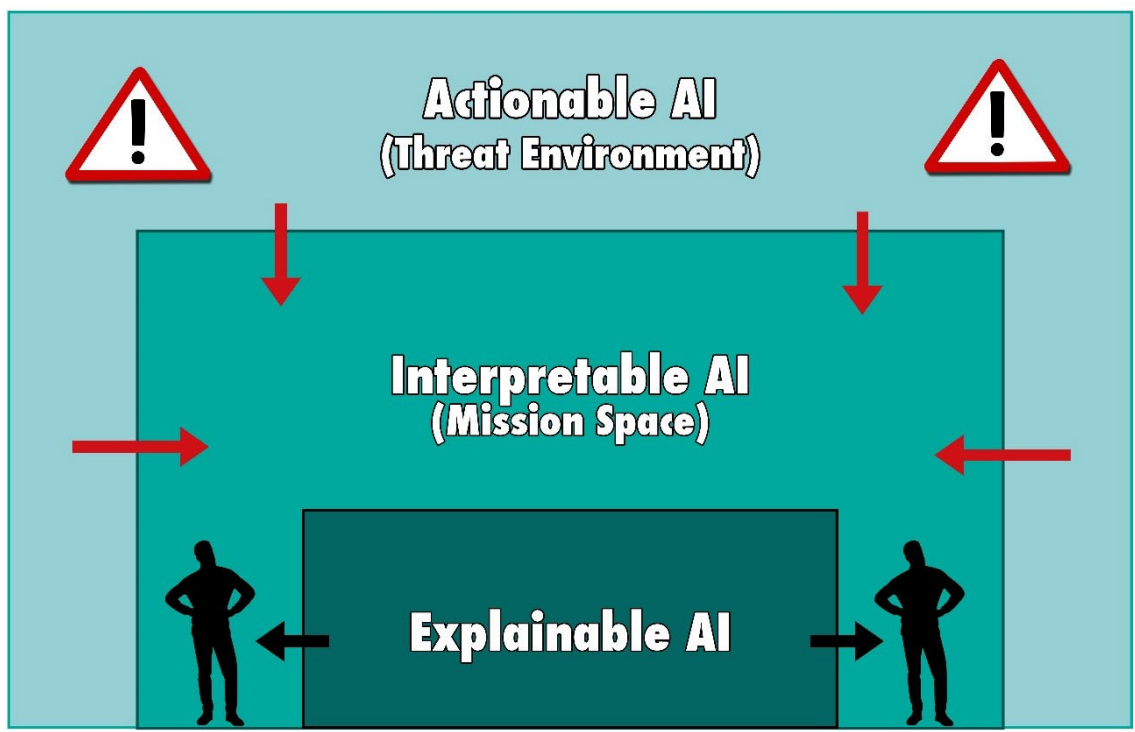

Figure 1. Explainable, Interpretable and Actionable AI.

AI systems should be able to capture values of decision maker in selecting courses of action, but also AI should provide level of confidence and sufficient information that the decision maker can critically evaluate its recommendation. Actionable Interpretable AI requires that operators understand enough about the decisions and their assumptions to anticipate how well-suited AI recommendations will be to the problem at hand. Thus, in addition to communicating reasoning processes, AI must be prepared to communicate important contextualizing factors to its users. Decision output should include projections of performance to various changes or challenges that may arise, according to the user objectives. AI

……………………………

output could also anticipate how those objectives might change, at least in framing, within different futures.

Ultimately, a near-term requirement to enhance AI includes deepening the contextualized interactions between AI and its users to build human trust in AI outputs. Actionable AI provides insight to AI decisions' value in uncertain futures, and Interpretable AI allows users to toggle parameter inputs to study the effects on the decisions. Together they enhance explainable AI as Actionable Interpretable AI ($AI^2$).

# Acknowledgement

We are grateful for the illustrations of George Siharulidze. The opinions expressed herein are those of the authors alone, and may not represent the opinions of their affiliate institutions.